%% file: neurips_2025.tex
\documentclass{article}

\usepackage[preprint]{neurips_2025}

\usepackage[utf8]{inputenc} 
\usepackage[T1]{fontenc}    
\usepackage{hyperref}       
\usepackage{url}            
\usepackage{booktabs}       
\usepackage{amsfonts}       
\usepackage{nicefrac}       
\usepackage{microtype}      
\usepackage{xcolor}         
\usepackage{amsmath}
\usepackage{graphicx}
\newcommand*{\method}{UniRL}
\usepackage{tabularx}
\usepackage{booktabs}
\definecolor{mygray}{gray}{0.6} 
\definecolor{lightblue}{RGB}{235, 245, 255}
\usepackage{colortbl}
\usepackage{wrapfig}
\usepackage{enumitem}
\usepackage{multirow}
\usepackage{marvosym}

\title{UniRL: Self-Improving Unified Multimodal Models via Supervised and Reinforcement Learning}

\author{
~Weijia Mao{$^{1}$ $^{2}$}, ~Zhenheng Yang{$^{2}$},~Mike Zheng Shou$^{1\textrm{\Letter}}$
\\\\\vspace{-8pt}
{$^{1}$}Show Lab, National University of Singapore {$^{2}$}ByteDance
}

\begin{document}

\renewcommand{\thefootnote}{\fnsymbol{footnote} }{}
\footnotetext{$^{\textrm{\Letter}}$ Corresponding Author}

\maketitle

\input{sections/abstract}

\input{sections/intro}

\input{sections/related_work}

\input{sections/methods}

\input{sections/experiments}
\input{sections/conclusion}

\bibliographystyle{plain}
\bibliography{neurips_2025}

\newpage
\appendix

\input{sup}

\end{document}

%% file: sections/abstract.tex
\begin{abstract}

Unified multimodal large language models such as Show-o and Janus have achieved strong performance across both generation and understanding tasks. However, these models typically rely on large-scale datasets and require substantial computation during the pretraining stage. In addition, several post-training methods have been proposed, but they often depend on external data or are limited to task-specific customization. In this work, we introduce \method{}, \textbf{a self-improving post-training approach}. Our approach enables the model to generate images from prompts and use them as training data in each iteration, without relying on any external image data. Moreover, it enables the two tasks to enhance each other: the generated images are used for understanding, and the understanding results are used to supervise generation. We explore supervised fine-tuning (SFT) and Group Relative Policy Optimization (GRPO) to optimize the models. \method{} offers three key advantages: (1) it requires no external image data, as all training samples are generated by the model itself during training; (2) it not only improves individual task performance, but also reduces the imbalance between generation and understanding; and (3) it requires only several additional training steps during the post-training stage. We evaluate \method{} on top of Show-o and Janus, achieving a GenEval score of 0.77 for Show-o and 0.65 for Janus. Code and models will be released in \url{https://github.com/showlab/UniRL}.

\end{abstract}

%% file: sections/intro.tex
\section{Introduction}

Unified multimodal large language models are designed to handle both generation and understanding tasks within a shared parameter space. Recently, models such as Show-o, Janus, and others~\cite{showo,janus,janusflow,transfusion,emu3} have made rapid progress by adopting various architectural and training strategies. Most existing work focuses on the pretraining stage, leveraging large-scale data to train models built upon large language models (LLMs)~\cite{phi1.5,phi3,llama} or multimodal LLMs (MLLMs)~\cite{qwen_vl,llava,llava-next}. However, these methods often face challenges such as high computational cost and the need for vast amounts of data. Meanwhile, several studies~\cite{hermesflow,showo_cot,mvot} have explored the post-training phase, but they face limitations: some require additional external data~\cite{hermesflow,dora_cycle,mvot}, others target only a single task~\cite{showo_cot,mvot}, or focus on task-specific customization~\cite{dora_cycle}. Therefore, developing effective post-training methods to improve the performance of unified multimodal models remains a significant challenge.

In this work, we propose \textbf{a self-improving post-training method} for unified multimodal models without external image data. We begin by constructing prompts and question–answer (QA) pairs inspired by the GenEval~\cite{geneval} benchmark, which evaluates generation quality by categorizing prompts into six types—such as counting, color, and position—designed to reflect fundamental visual features of natural images. We adopt this categorization to design the prompts used in our training process. In each training iteration, a constructed prompt is fed into the model to generate a group of images. These images, along with the corresponding questions, are then input back into the model to predict answers. The model is optimized using both the predicted and ground-truth answers to improve generation and understanding simultaneously. For optimization, we explore two strategies: supervised fine-tuning (SFT) and reinforcement learning.

In the post-training stage, supervised fine-tuning (SFT) and reinforcement learning have traditionally been widely used. Recently, Group Relative Policy Optimization (GRPO)~\cite{grpo}, a reinforcement learning method, has shown strong performance in large language models (LLMs)~\cite{deepseek_r1}, enabling effective post-training optimization through chain-of-thought (CoT) reasoning. In this work, we explore two optimization strategies—SFT and GRPO—for unified multimodal models and analyze the advantages of each strategy. Previous applications of GRPO typically rely on chain-of-thought outputs to estimate reward distributions. However, most current unified multimodal models lack the ability to produce structured reasoning steps. Therefore, we explore a GRPO-based method that does not rely on reasoning outputs, aiming to improve the performance of unified multimodal models.

\begin{figure*}[t]
    \centering
    \includegraphics[width=1\linewidth]{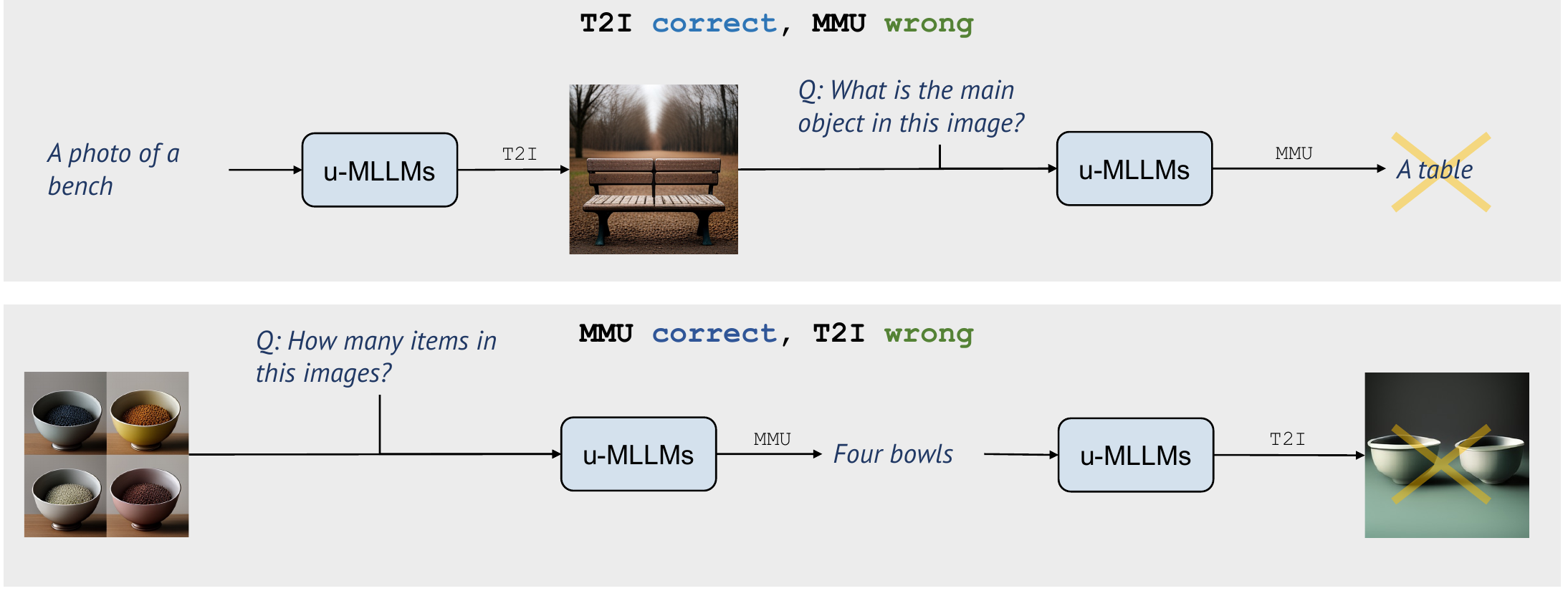}
    \vspace{-15pt}
    \caption{The imbalance between Text-to-Image Generation (T2I) and Multi-modal Understanding (MMU). For the same image, unified multimodal models may struggle to perform both generation and understanding consistently.}
    \label{fig:teaser}
    \vspace{-15pt}
\end{figure*}


For the performance of the unified multimodal models, we define it along three dimensions: (1) image generation capability, (2) image understanding capability, and (3) the balance between the two.  While prior work has primarily focused on improving performance for a single task, unified multimodal models frequently exhibit poor alignment between the two capabilities. As shown in Fig.~\ref{fig:teaser}, these models often fail to simultaneously support both tasks on the same image. The model may succeed in generating a plausible image but fail to answer the corresponding question, or fail in generation while answering correctly. Therefore, beyond improving individual tasks, our self-improving method explicitly targets the imbalance between image generation and understanding. To quantitatively assess this imbalance, we design a new evaluation metric. While visual understanding spans a wide range of tasks, including reasoning and mathematical problem solving, we focus on basic visual features of natural images—such as object count, color, and position—in both training and evaluation to ensure alignment between generation and understanding tasks.

Our method offers three main advantages. (1) It does not rely on external image data, using only images generated in real time during training. (2) It improves the performance of individual tasks while also reducing the imbalance between generation and understanding. Specifically, the same set of generated images is used for both tasks, enabling shared supervision that helps improve the imbalance between them. (3) It requires only a small number of training steps during the post-training stage. To evaluate our method, we adopt Show-o and Janus as base models, assessing improvements in both single-task performance and the balance between tasks. Our approach achieves post-training GenEval scores of 0.77 on Show-o and 0.65 on Janus.

Our main contributions are summarized as follows:
    


    

\begin{itemize}[itemsep=0em, topsep=0em]
    \item To the best of our knowledge, we propose the first self-improving post-training optimization method for unified multimodal models via supervised and reinforcement learning.
    
    \item We investigate the effectiveness of both supervised fine-tuning (SFT) and Group Relative Policy Optimization (GRPO) for unified multimodal models, and provide a detailed comparison of their respective performance.

    \item We introduce a new metric to quantify the imbalance between generation and understanding tasks within a single model. Our evaluations across multiple unified models reveal that such imbalance is common.

    \item Our method substantially improves image generation and understanding quality and effectively reduces task imbalance, enabling better consistency between text-to-image generation and multimodal understanding.
\end{itemize}

%% file: sections/related_work.tex
\section{Related work}
\subsection{Unified multimodal understanding and generation}
Several recent studies~\cite{seed-x,wu2023next,team2024chameleon,showo,lwm,sun2023emu,transfusion,CoDI,dreamllm,emu3,janusflow,lumina_mgpt,liquid,llamafusion,anil2023gemini,tokenflow,synergen,Orthus,omniflow,januspro,metamorph} have explored unified transformers capable of both generation and comprehension. Chameleon~\cite{team2024chameleon},Emu3~\cite{emu3}, Janus~\cite{janus}, JanusPro~\cite{januspro} adopt autoregressive methods for both tasks. SEED-X~\cite{seed-x} incorporates a diffusion model alongside a large language model to support multimodal generation and understanding. While both Show-o~\cite{showo} and JanusFlow~\cite{janusflow} use autoregression for understanding, the former adopts discrete diffusion for generation, whereas the latter employs flow matching. These models primarily focus on the pretraining stage using different methods for different tasks. These approaches require substantial training time and large-scale training data. Some studies~\cite{Unimod,showo_cot} have explored more efficient training or inference strategies. In this work, we take Show-o and Janus as our base model and explore post-training optimization techniques to further improve unified multimodal performance.

\subsection{GRPO in LLM and MLLMs}
Supervised fine-tuning (SFT) and reinforcement learning have been widely adopted in the post-training of large language models (LLMs)~\cite{gpt_4o,phi1.5,phi3,llama} and multimodal LLMs (MLLMs)~\cite{qwen_vl,llava}. Group Relative Policy Optimization (GRPO) is introduced in~\cite{grpo} to optimize LLMs by generating a group of chain-of-thought outputs and applying a task-specific reward function to guide learning. More recently, GRPO has also been explored in the context of MLLMs~\cite{grpo_vl}. In addition to GRPO, other reinforcement learning approaches such as Direct Preference Optimization (DPO)~\cite{dpo,hermesflow} and Proximal Policy Optimization (PPO)~\cite{ppo} have also been applied to both LLMs and MLLMs. Hermesflow~\cite{hermesflow} and Emu3~\cite{emu3} use DPO to optimize unified multimodal models. However, PPO requires a separate value network to estimate the baseline, and DPO relies on predefined positive and negative examples, which introduces additional complexity. In our work, we explore SFT and GRPO method to optimize unified multimodal models and improve their generation and understanding performance during the post-training stage.

%% file: sections/methods.tex
\section{Method}
In this section, we first introduce the background of unified multimodal models and the GRPO optimization method (Sec.\ref{preliminary}). We then describe the construction of training prompts (Sec.\ref{prompt_construction}), the proposed self-improving framework (Sec.\ref{self-evolve}), and two optimization strategies: SFT and GRPO (Sec.\ref{sft_grpo_optimization}). We also present a non-end-to-end training approach designed for models that use different image representations in the two tasks (Sec.\ref{end-to-end-training}). Finally, we introduce a new metric to evaluate the imbalance between the two tasks (Sec.\ref{new_metric}). The pipeline of training is shown in Fig.~\ref{fig:pipeline}.

\subsection{Preliminary}
\label{preliminary}

\textbf{Unified Multimodal Models.} Unified multimodal models can be divided into two main categories: (1) both generation and understanding tasks use autoregressive methods; (2) generation makes use of diffusion or flow matching while comprehension relies on autoregressive methods. We focus on two example models: \textbf{Show-o}~\cite{showo} and \textbf{Janus}~\cite{janus}.


In Show-o, the text tokens are denoted as \(T = \{t_1, \dots, t_N\}\) and the image tokens as \(I = \{i_1, \dots, i_M\}\). The training objective includes two components: one maximizes the likelihood of each text token given all previous text and image tokens; the other reconstructs masked image tokens based on the remaining image tokens and the full text sequence. The two loss terms are:

\begin{equation}
\mathcal{L}_{\mathrm{NTP}}
= \sum_{n=1}^{N}\log p_{\theta}\bigl(t_{n}\mid t_{1:n-1},\,i_{1:M}\bigr),\quad
\mathcal{L}_{\mathrm{MTP}}
= \sum_{m=1}^{M}\log p_{\theta}\bigl(i_{m}^{*}\mid i_{1:m-1},\,i_{m+1:M},\,t_{1:N}\bigr).
\end{equation}

Janus uses only the NTP objective, predicting every next token (image or text) in a single autoregressive pass.

\textbf{Group Relative Policy Optimization(GRPO)} 
We use a reinforcement learning method called Group Relative Policy Optimization (GRPO)~\cite{grpo}, which optimizes the model by comparing a group of generated outputs and assigning relative rewards. Suppose inputs \(x\) come from a distribution \(D\). For each \(x\sim D\), we sample \(M\) candidate outputs \(\{y_{j}\}_{j=1}^{M}\) under an old policy \(\pi_{\varphi_{\mathrm{old}}}\). We then update our new policy \(\pi_{\varphi}\) by maximising the following objective:

\begin{equation}
\label{eq:grp_update}
L(\varphi) = 
\mathbb{E}_{\substack{x \sim D,\\ \{y_{j}\}\sim \pi_{\varphi_{\mathrm{old}}}}}
\Biggl[
\frac{1}{M}
\sum_{j=1}^{M}
\min\bigl(s_{j}\,A_{j},\,\mathrm{clip}(s_{j},\,1-\delta,\,1+\delta)\,Z_{j}\bigr)
\Biggr]
- \lambda\,D_{\mathrm{KL}}\!\bigl(\pi_{\varphi}\,\|\;\pi_{\psi}\bigr),
\end{equation}

where
\[
s_{j} \;=\;
\frac{\pi_{\varphi}(y_{j}\mid x)}
{\pi_{\varphi_{\mathrm{old}}}(y_{j}\mid x)},
\quad
A_{j} \;=\;
\frac{R_{j} \;-\;\mathrm{mean}(R_{1},\dots,R_{M})}
{\mathrm{std}(R_{1},\dots,R_{M})}.
\]
\(R_{j}\) is the reward for output \(y_{j}\), \(\delta\) sets the clipping range, \(\lambda\) weighs the divergence penalty, and \(\pi_{\psi}\) is a fixed reference policy.
In the inner sum, \(s_{j}\) measures how the new policy probability differs from the old policy for the same output. Multiplying by the normalised advantage \(A_{j}\) gives a gradient direction. Clipping \(s_{j}\) to \([1-\delta,\,1+\delta]\) ensures that no single sample drives the update too far. Averaging over \(M\) outputs makes the estimate stable across different group sizes. Finally, subtracting \(\lambda\,D_{\mathrm{KL}}(\pi_{\varphi}\,\|\,\pi_{\psi})\) discourages large shifts away from a trusted policy, which helps to keep learning smooth.

\input{tables/prompt}

\subsection{Prompt and QA pair construction}
\label{prompt_construction}

In our study, we are inspired by the GenEval~\cite{geneval} benchmark to construct prompt and question–answer pairs. The GenEval benchmark evaluates image generation models by dividing spatial relationships into six categories: \textit{single object}, \textit{two objects}, \textit{counting}, \textit{colors}, \textit{color attributes}, and \textit{position}. We create prompts and matching question–answer pairs based on these categories. First, beyond the eighty object classes in the COCO dataset, we use GPT-4o to generate eighty more everyday objects, resulting in 160 objects in total. Second, we follow the GenEval procedure to generate prompts and then formulate the corresponding questions and answers. For example, in the counting category, we use a prompt such as “a photo of three vases”, pair it with the question “How many items are in this image?”, and assign “three vases” as the answer. To ensure a strict separation between training and evaluation, we only use prompts from GenEval for testing. All prompts used for training are independently constructed and do not overlap with those in the GenEval benchmark. Additional examples appear in Tab.~\ref{tab:category_prompt_question_answer}. 

\begin{figure*}[t]
  \centering
  \includegraphics[width=\linewidth]{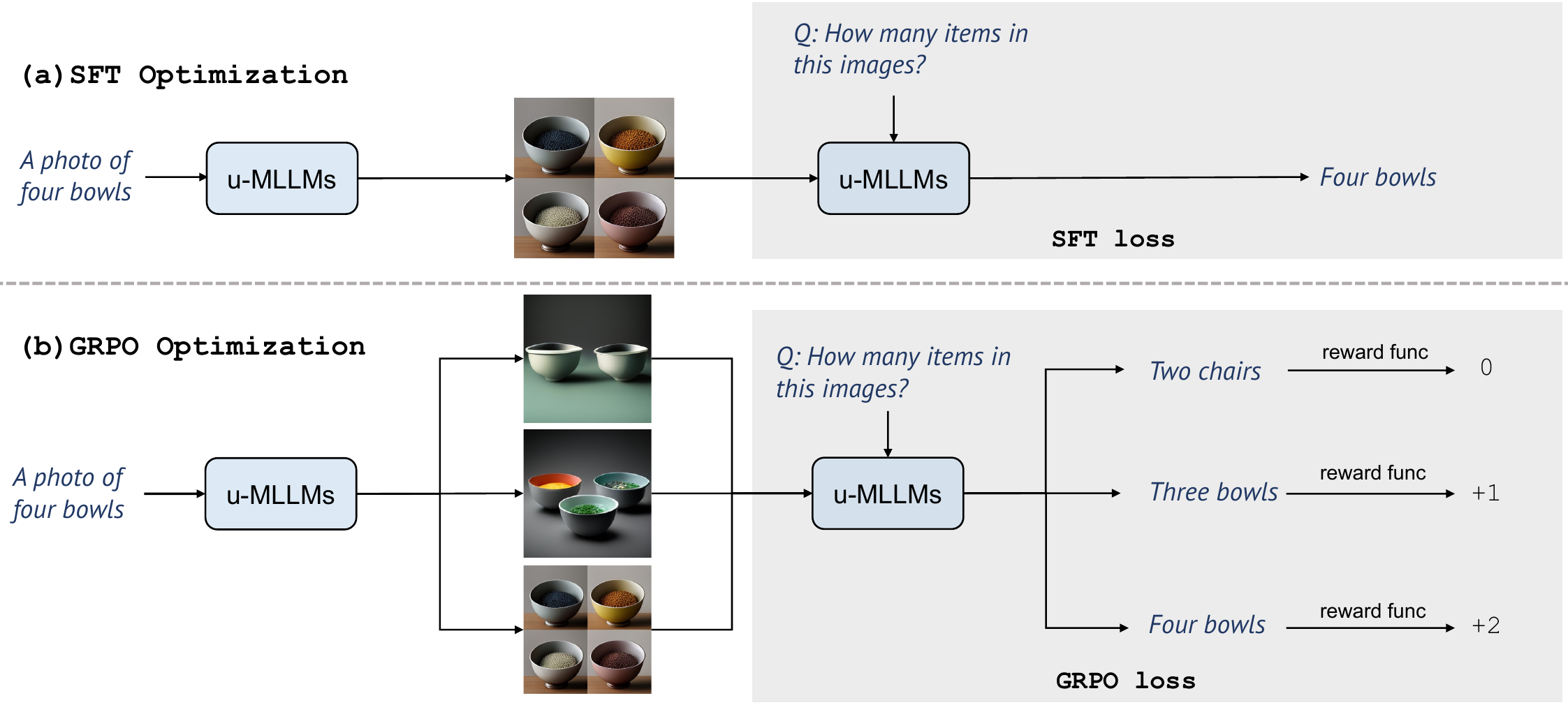}
  \vspace{-15pt}      

  \caption{Training pipeline of \method{} (a) SFT Optimization: A prompt is used to generate a single image, which, together with the corresponding question, is used to predict an answer and compute the SFT loss. (b) GRPO Optimization: The same prompt generates a set of images, each paired with the same question to produce multiple answers, which are used to compute the GRPO loss.}

  \label{fig:pipeline}
  \vspace{-15pt}      
\end{figure*}

\subsection{Self-improving optimization}
\label{self-evolve}

We propose a self-improving training framework that requires no external image data. In this framework, the image generation and multimodal understanding tasks mutually reinforce each other: the understanding task provides feedback to evaluate and guide image quality, while the generation task supplies training data that enhances the model’s understanding capability.

As shown in Fig.~\ref{fig:pipeline}, given a constructed prompt, the model first generates a sequence of image tokens \(\mathbf{u} = \{u_1, \dots, u_M\}\). Regardless of whether the image generation module uses a diffusion-based or autoregressive pipeline, we feed the generated image tokens and the prompt back into the model following the original T2I training process to obtain the image logits. Let \(\mathbf{z} = \{z_1, \dots, z_L\}\) denote the output logits corresponding to the \(L\) target positions to be predicted.

To enable gradient flow through the discrete image tokens, we apply the Straight-Through Gumbel-Softmax (ST-GS) estimator to obtain differentiable approximations:

\begin{equation}
\hat{u}_l = \mathrm{ST\text{-}GumbelSoftmax}(z_l), \quad \text{for } l = 1, \dots, L,
\end{equation}

resulting in a completed image token sequence \(\hat{\mathbf{u}}\) that includes differentiably sampled tokens. The formulation and implementation details of the Straight-Through Gumbel-Softmax estimator are provided in Sec.~\ref{gumbel_softmax_formula}. This image sequence, together with the corresponding constructed question \(q\), is then used to predict an answer. We use the predicted answers and ground-truth answer to optimize the model exploring two techniques, SFT and GRPO.

\subsection{SFT and GRPO optimization}
\label{sft_grpo_optimization}

\textbf{SFT optimization.} We use the predicted answers and the ground-truth answers to calculate the SFT loss to optimize the models. The SFT loss is computed as:

\begin{equation}
\mathcal{L}_{\text{SFT}} = -\sum_{t=1}^{T} \log p_\theta(y_t \mid y_{<t}, \hat{\mathbf{u}}, q),
\end{equation}

where \(y_t\) is the \(t\)-th token of the ground-truth answer, and \(y_{<t}\) are the preceding tokens. This formulation applies generally to both masked and autoregressive image generation models. The variable \(\hat{\mathbf{u}}\) denotes the generated image token sequence, which includes differentiably sampled tokens and serves as input to the answer generation module. This formulation is compatible with both masked and autoregressive image generation models. The SFT loss enables joint optimization of image generation and understanding tasks by allowing gradients to flow through the image tokens.

\textbf{GRPO optimization.} We also adopt Group Relative Policy Optimization (GRPO) to optimize the unified multimodal models. During the GRPO phase, the model takes a prompt \(p\) and generates a group of \(K\) images \(\{\hat{\mathbf{u}}_1, \dots, \hat{\mathbf{u}}_K\}\). Each image \(\hat{\mathbf{u}}_k\), together with its corresponding question \(q\), is then fed back into the model to produce a predicted answer \(\hat{a}_k\). A reward \(r_k = r(\hat{a}_k)\) is computed by comparing the predicted answer to the ground-truth answer \(a^*\), based on the designed reward function. Similar to SFT, we also apply the Straight-Through Gumbel-Softmax estimator to enable differentiable sampling of image tokens, allowing gradients to propagate through both generation and understanding tasks. The overall GRPO loss is defined as:

\begin{equation}
\mathcal{L}(\theta)
=
\mathbb{E}_{\substack{x \sim D,\\ \{\hat{a}_k\} \sim \pi_{\theta_{\mathrm{old}}}}}
\left[
-\,\sum_{k=1}^{K}
w_k\,\log p_\theta\bigl(\hat{a}_k \mid \hat{\mathbf{u}}_k,\,q\bigr)
\right]
+
\beta\,
\mathrm{KL}\!\left(p_\theta(a \mid \hat{\mathbf{u}}, q)\,\|\,p_{\theta_{\mathrm{ref}}}(a \mid \hat{\mathbf{u}}, q)\right),
\end{equation}

where the importance weight \(w_k\) is computed as:

\[
w_k = \frac{\exp\left(\alpha\,(r_k - r_{\mathrm{mean}})\right)}{\sum_{j=1}^{K} \exp\left(\alpha\,(r_j - r_{\mathrm{mean}})\right)},
\]

with \(r_{\mathrm{mean}}\) denoting the average reward within the group and \(\alpha\) controlling the sharpness of the weight distribution. 


The first term in the loss is the GRPO objective, which applies a weighted log-likelihood over the \(K\) sampled answers. \(\pi_{\theta_{\mathrm{old}}}\) is the fixed sampling policy, typically a frozen copy of the model, while \(\pi_\theta\) (or \(p_\theta\)) is the current policy being optimized. Rewards are centered by subtracting the group mean \(r_{\mathrm{mean}}\), scaled by temperature \(\alpha\), and converted into weights \(w_k\) to emphasize better answers. The second term is a KL penalty, scaled by \(\beta\), to keep the updated policy close to a reference model, improving stability and preventing mode collapse.

\input{tables/reward}

Importantly, because the image tokens \(\hat{\mathbf{u}}_k\) are sampled in a differentiable manner, the reward signals from answer prediction can flow through both the understanding and image generation branches. This enables effective end-to-end optimization of both tasks.

\textbf{Reward Function.}  
The reward function assigns scores based on alignment between the predicted answer \(\hat{a}_k\) and the ground-truth answer \(a^*\). While the specific reward criteria vary by question category, all rewards are based on two main components:  
(1) correct identification of object names, and  
(2) accurate prediction of associated attributes, such as number, color, or spatial position.  
The detailed reward rules for each category are summarized in Tab.~\ref{reward}.

\subsection{End-to-end training vs. non-end-to-end training}
\label{end-to-end-training}


Unified multimodal models such as Show-o~\cite{showo}, Emu3~\cite{emu3}, Chameleon~\cite{team2024chameleon}, and Vila-u~\cite{vila-u} use shared image representations—typically via VQGAN~\cite{vqgan} or custom tokenizers—allowing end-to-end training. Gradients from MMU loss can flow through the T2I module via the Straight-Through Gumbel-Softmax estimator. In contrast, Janus~\cite{janus} and JanusPro~\cite{januspro} use separate representations: discrete tokens for T2I and ViT features~\cite{vit} for MMU. This mismatch prevents effective gradient flow; although gradients can technically propagate, they cannot meaningfully guide the discrete tokens, often leading to unstable training.

To handle such cases, we adopt a non-end-to-end training strategy. For T2I, rewards computed from MMU performance are used to directly optimize the image generation module. Conversely, for MMU, the model generates images, answers the corresponding questions, and uses the rewards to update the MMU branch. In the end-to-end setting, a unified GRPO loss is used to jointly optimize both tasks. In contrast, the non-end-to-end setting applies separate GRPO objectives to T2I and MMU. Although the two modules are trained independently, reward signals still enable cross-task interaction, mitigating the limitations of mismatched representations. Full details and loss formulas are provided in the Sec.~\ref{non_end_to_end}.

\subsection{The balance between image understanding and generation}
\label{new_metric}

\input{tables/metric_model}

To evaluate the balance between text-to-image generation (T2I) and multimodal understanding (MMU), we propose a bidirectional evaluation metric with two chains: T2I→MMU and MMU→T2I. In T2I→MMU, the model generates an image from a prompt and answers a question only if the image is correct; we then compute the answer accuracy. In MMU→T2I, the model first answers a question based on a given image; if correct, we regenerate the image from the prompt and evaluate its quality. This setup assesses the consistency between generation and understanding. The conditional accuracies are defined as:

\begin{equation}
\text{Accuracy}_{\text{MMU}|\text{T2I}} = \frac{N_{\text{A} \cap \text{I}}}{N_{\text{I}}}, \quad
\text{Accuracy}_{\text{T2I}|\text{MMU}} = \frac{N_{\text{A} \cap \text{I}}}{N_{\text{A}}}.
\end{equation}

\(N_{\text{I}}\) denotes the number of correctly generated images, \(N_{\text{A}}\) the number of correctly answered questions, and \(N_{\text{A} \cap \text{I}}\) the number of samples where both are correct. These metrics quantitatively assess the consistency between generation and understanding. \(\text{Accuracy}_{\text{MMU}|\text{T2I}}\) measures answer correctness given a correct image, while \(\text{Accuracy}_{\text{T2I}|\text{MMU}}\) measures image correctness given a correct answer. This bidirectional formulation evaluates the alignment between the two tasks beyond isolated performance. As shown in Tab.~\ref{tab:metric_model}, we apply this metric to several unified models, including Show-o~\cite{showo}, Janus~\cite{janus}, Janus-Pro~\cite{januspro}, and Vila-u~\cite{vila-u}, and observe that task imbalance is consistently present across all models. More details can be seen in Sec.~\ref{more_implementation_detail}.

%% file: tables/prompt.tex
\begin{table}[htbp]
  \centering
  \label{tab:category_prompt_question_answer}
  \vspace{-5pt}
  \caption{Examples of prompts constructed for six categories of image-based question–answer tasks: single object, two object, counting, colors, position, and attribute.}

  \small
   \begin{tabular}{
    l
    >{\raggedright\arraybackslash}p{0.20\linewidth}
    >{\raggedright\arraybackslash}p{0.35\linewidth}
    >{\raggedright\arraybackslash}p{0.2\linewidth}
  }
    \toprule
    Category   & Prompt                                                            & Question & Answer \\
    \midrule
    Single object
      & a photo of a bench
      & What is the main object of the image?       &  Bench     \\
    Two object
      & a photo of a table and a soccer
      & What are two main objects of the image?        & A table and a soccer    \\
    Counting
      & a photo of three vases
      & How many items in this image?        & Three vases      \\
    Colors
      & a photo of a green fork
      & What is the color of the object?        & Green      \\
    Position
      & a photo of a train above of an elephant
      & What are two objects and what is position relationship between two main items?        & The train is above of an elephant      \\
    Attribute
      & a photo of a blue chair and a red umbrella
      & What are two objects and the colors of two objects in the image?         & The chair is  blue and  the umbrella is red    \\
    \bottomrule
  \end{tabular}
  \vspace{-15pt}
\end{table}

%% file: tables/reward.tex

\begin{wraptable}{r}{0.4\textwidth}
  \centering
  \setlength{\tabcolsep}{1.8pt}   
  \small                       
  \vspace{-10pt}
  \caption{Examples of reward rules based on key words in generated answers}
  \label{reward}
  \begin{tabular}{l c >{\raggedright\arraybackslash}p{0.18\textwidth}}
    \toprule
    Category      & Score & Key Word                  \\
    \midrule
    Single object & 1     & bench                     \\
    Two object    & 2     & table, soccer       \\
    Counting      & 2     & three, vase          \\
    Colors        & 1     & green                      \\
    Position      & 3     & train, above of, elephant \\
    Attribute     & 4     & blue, chair,\\
                  &       & red, umbrella          \\
    \bottomrule
    \vspace{-25pt}
  \end{tabular}
\end{wraptable}

%% file: tables/metric_model.tex
\begin{wraptable}{r}{0.35\textwidth}
  \centering
  \vspace{-15pt}
  \caption{Performance of the proposed metric for evaluating the imbalance between two tasks.}
  \label{tab:metric_model}
   \setlength{\tabcolsep}{1pt}
  \resizebox{\linewidth}{!}{%
    \begin{tabular}{lcc}
      \toprule
      Method      & $\text{Accuracy}_{\text{MMU}|\text{T2I}}$   & $\text{Accuracy}_{\text{T2I}|\text{MMU}}$ \\
      \midrule
      Show-o   & 0.83        & 0.70              \\
      Janus    & 0.67        & 0.68              \\
      JanusPro & 0.69        & 0.85              \\
      Vila-u   & 0.68        & 0.85              \\
      \bottomrule
    \end{tabular}%
  }
  \vspace{-15pt}
\end{wraptable}

%% file: sections/experiments.tex
\section{Experiments}

\subsection{Implementation details}
\label{implementation_details}


We evaluate our method on two base models: Show-o~\cite{showo} and Janus~\cite{janus}. For \textbf{Show-o}, we adopt end-to-end training. The SFT stage uses a batch size of 2 for 5,000 iterations, and the GRPO stage is configured with a group size of 3, a KL coefficient \(\beta = 0.2\), and 3,000 training steps on 8 H100 GPUs. During inference, we set the guidance scale to 2, diffusion steps to 16, and resolution to 512. For \textbf{Janus}, due to its use of different representations for generation and understanding, we apply a non-end-to-end training strategy. The model is trained with a batch size of 8 for 1,000 iterations using the same KL coefficient. Inference is performed with a guidance scale of 5 and resolution of 384. More implementation details can be seen in Sec.~\ref{more_implementation_detail}.

 \subsection{Quantitative results}
\label{quantitative_results}

\textbf{Baselines and our method.}  
We compare our method against several baselines.  \textit{Original Model.} We evaluate the performance of the pretrained model without any post-training.   \textit{HermesFlow.}~\cite{hermesflow} This method fine-tunes the Show-o model with DPO optimization on the JourneyDB dataset~\cite{journeydb}. We evaluate two variants of our post-training framework:  
\textit{\method{} (SFT).} Our method using supervised fine-tuning (SFT) without reinforcement learning. \textit{\method{} (GRPO).} Our method using Group Relative Policy Optimization (GRPO). For the Janus model, we only compare our GRPO-based method with the original model, as applying SFT leads to unstable training and failure. The specific reasons are discussed in Sec.~\ref{analysis}.


\textbf{Text to image generation.} For the image generation task, we evaluate model performance using three benchmarks: GenEval~\cite{geneval}, DSG-1K~\cite{DSG}, and CLIP-Score~\cite{clip}. To calculate the CLIP score, we use 2,000 generated images sampled from the GenEval and DSG-1K benchmarks. As shown in Tab.~\ref{tab:t2i}, our method with SFT improves the GenEval score from 0.60 to 0.76, while the GRPO-based variant achieves 0.71—both outperforming the baselines by a large margin. In addition, our approach yields notable gains on DSG-1K and CLIP-Score, demonstrating the effectiveness of our post-training strategy in improving image generation quality. More results will be shown in Sec.~\ref{geneval_more}.

\input{tables/geneval}

\textbf{Multimodal understanding.}  
We evaluate the model's ability to understand natural visual features across six categories: \textit{single object}, \textit{two objects}, \textit{counting}, \textit{colors}, \textit{position}, and \textit{color attributes}. These categories capture core aspects of visual understanding, including object recognition, counting, color perception, and spatial reasoning—fundamental properties of natural images. To assess generalization beyond the training distribution, we use images generated by external models such as JanusPro. The generated images, along with their corresponding questions, are then fed into the target model to evaluate its understanding capabilities on out-of-distribution visual inputs. We compute the accuracy for each categories, allowing a fine-grained analysis of the model’s generalization across different aspects of visual understanding.

As shown in Tab.~\ref{tab:mmu}, our SFT-based method performs slightly worse than the original model, primarily due to overfitting, particularly in the single-object category. In contrast, our GRPO-based method achieves an overall score of 0.79, significantly outperforming all baselines. The underlying reasons are discussed in Sec.~\ref{analysis}, while additional evaluation details and benchmark results are provided in Sec.~\ref{more_implementation_detail} and Sec.~\ref{more_mmu_result}.

\input{tables/mmu}
\textbf{The imbalance between two tasks.}  
To evaluate the balance between image understanding and generation, we apply our proposed metric, as shown in Tab.~\ref{tab:mmu_t2i}. The prompts used for \(\text{Accuracy}_{\text{MMU}|\text{T2I}}\) are taken from the GenEval benchmark, while those for \(\text{Accuracy}_{\text{T2I}|\text{MMU}}\) are aligned with the MMU evaluation setup. The results show that our GRPO-based method significantly reduces the imbalance between the two tasks. For certain subtasks, the model achieves nearly 100\% accuracy, demonstrating strong consistency between generation and understanding capabilities.

\input{tables/t2i_mmu}
\input{tables/mmu_t2i}




\subsection{Visualization results}

\begin{figure*}[t]
  \centering
  \includegraphics[width=\linewidth]{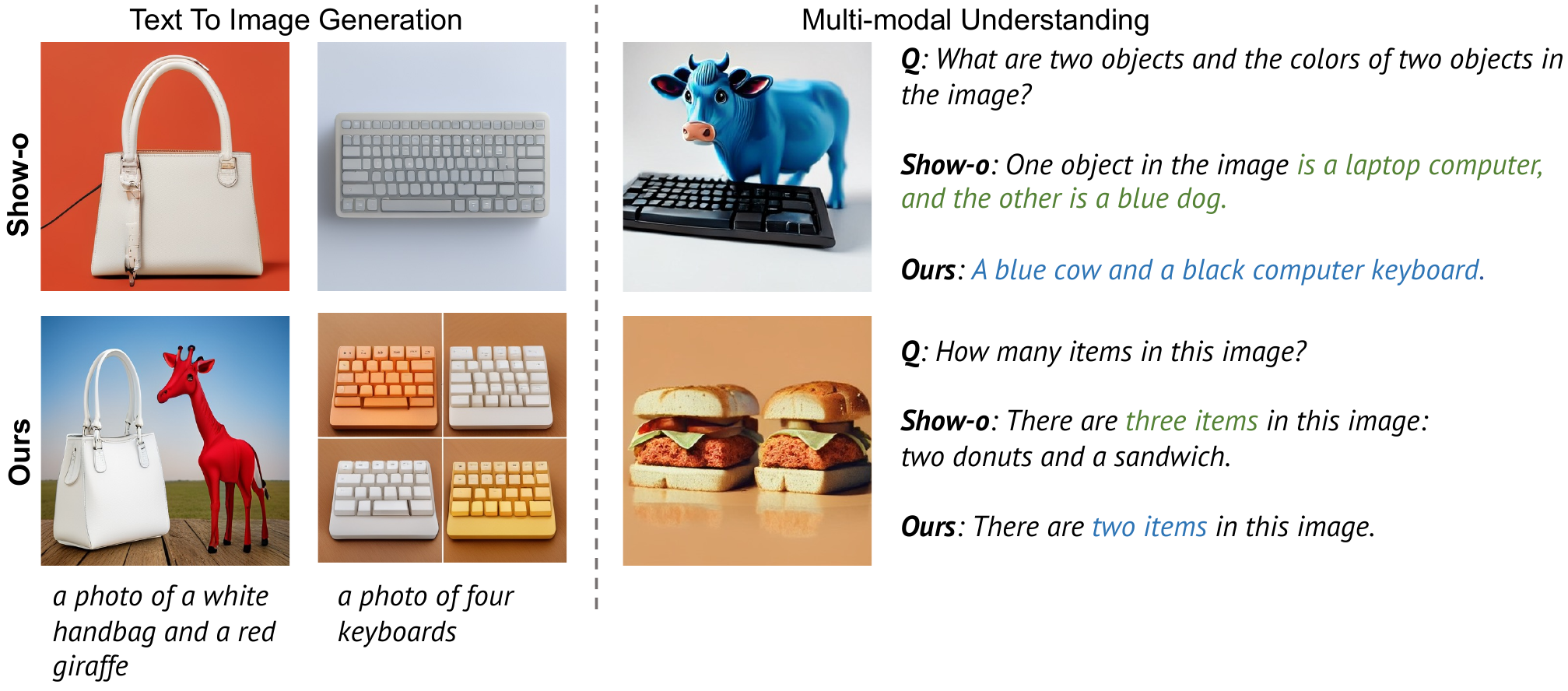}
  \vspace{-15pt}
  \caption{Qualitative comparison of our method with the original models on both text-to-image generation (T2I) and multimodal understanding (MMU) tasks.}
  \label{fig:result}
  \vspace{-15pt}
\end{figure*}

As shown in Fig.~\ref{fig:result}, we present qualitative comparisons between our method and the original models on both tasks. Our approach produces more visually faithful generations conditioned on prompts and yields more accurate answers for corresponding questions, demonstrating improved consistency across the two tasks. More results will be shown in Sec.~\ref{ood_result}.

\subsection{Analysis of SFT vs. GRPO}
\label{analysis}

From these experimental results, we can draw some key observations.

\textbf{Observation 1}: \textit{SFT is more prone to overfitting and memorization than GRPO in the image understanding task.}

As shown in Tab.~\ref{tab:mmu}, SFT performs significantly worse than both the original model and GRPO on the \textit{single object} category, where the training and test sets are fully disjoint in both examples and object types. In contrast, categories like \textit{position} and \textit{colors} share abstract features (e.g., position or color attributes), which partially mitigate overfitting. Compared to SFT, GRPO demonstrates better generalization under the same conditions.

\textbf{Observation 2}: \textit{In the generation task, SFT outperforms GRPO under an end-to-end training pipeline.}

As shown in Tab.~\ref{tab:t2i}, SFT achieves better T2I results than GRPO, with no noticeable overfitting. We hypothesize two reasons for this difference. (1) MMU is more prone to overfitting than T2I, as it mainly learns from text token patterns, while T2I involves more structured image reconstruction. (2) Our training pipeline is end-to-end. The SFT supervision signal for MMU is more direct and localized, whereas for T2I it is more indirect and distributed, which may promote better generalization in image generation. Under indirect supervision in T2I, SFT may optimize more effectively, as its loss is directly based on ground-truth answers. More analysis will be discussed in Sec.~\ref{analysis_non_end_to_end_sft}.

\textbf{Observation 3}: \textit{For non-end-to-end training setups such as Janus, SFT leads to failure in the image generation task.}  

If we directly use the model’s generated images to supervise T2I at each iteration, training will become unstable. We speculate that early-stage generation errors are repeatedly reinforced, leading to input distribution drift and eventual collapse. Therefore, for unified models with different representations for the two tasks, only the GRPO-based method is effective for improving performance. More analysis will be discussed in Sec.~\ref{analysis_non_end_to_end_sft}.

Overall, GRPO is a more general and robust optimization strategy compared to SFT. \textbf{Regardless of the specific task or model architecture—whether the two tasks share the same image representation or not—GRPO consistently improves performance and effectively reduces the imbalance between generation and understanding.} In contrast, while SFT tends to perform better on tasks like T2I in end-to-end setups (e.g., Show-o), it is more prone to memorization and overfitting.


%% file: tables/geneval.tex
      

\begin{table*}[t]
  \centering
  \caption{Comparison of \method{} with baseline methods across Text-to-Image (T2I) benchmarks.}
  \label{tab:t2i}
  \resizebox{\linewidth}{!}{%
    \begin{tabular}{lcccccc|ccc}  
      \toprule
      \multirow{2}{*}{Method}
        & \multicolumn{7}{c}{GenEval $\uparrow$}
        & \multirow{2}{*}{DSG-1K $\uparrow$}
        & \multirow{2}{*}{Clip score $\uparrow$} \\
      \cmidrule(lr){2-8}
        & Single. & Two. & Count. & Colors & Posi.& Attri.& Overall &  &  \\
      \midrule 
      Show-o     & 0.96 & 0.64 & 0.67 & 0.81 & 0.25 & 0.37 & 0.60 & 0.77 & 0.331 \\
      
      HermersFlow     & 0.97 & 0.67 & 0.65 & 0.77 & 0.28 & 0.42 & 0.61 & 0.78 & 0.334 \\
      \method{}(SFT)     & 0.99 & 0.93 & 0.62 & 0.89 & 0.55 & 0.68 & 0.77 & 0.82 & 0.337 \\
      \rowcolor{lightblue}
      \method{}(GRPO)  & 0.95 & 0.77 & 0.65 & 0.82 & 0.50 & 0.60 & 0.71 & 0.80 & 0.335 \\
      \midrule 
       Janus  & 0.96 & 0.64 & 0.29 & 0.81 & 0.49 & 0.46 & 0.60 & 0.80 & 0.332 \\
       \rowcolor{lightblue}
       \method{}(GRPO)  & 0.95 & 0.74 & 0.27 & 0.81 & 0.62 & 0.52 & 0.65 & 0.81 & 0.335 \\
      \bottomrule
    \end{tabular}%
  }
  \vspace{-15pt}
\end{table*}

%% file: tables/mmu.tex

\begin{wraptable}{r}{0.55\textwidth}
  \centering
  \vspace{-15pt}
    \caption{Comparison of \method{} with baseline methods across Multi-modal Understanding(MMU) benchmarks.}
  \label{tab:mmu}
  \setlength{\tabcolsep}{1.8pt}  
  \resizebox{\linewidth}{!}{%
    \begin{tabular}{lcccccc|c}
      \toprule
      Method           & Single. & Two. & Count. & Colors & Posi. & Attri. & Overall$\uparrow$ \\
      \midrule
      Show-o   & 0.99        & 0.88     & 0.79     & 0.98   & 0.20     & 0.54         & 0.71              \\
      HermesFlow & 0.99        & 0.88     & 0.79     & 0.98   & 0.34     & 0.52         & 0.73              \\
      \method{}(SFT)
      & 0.08        & 0.87     & 0.71     & 0.99   & 0.45     & 0.85        & 0.67              \\
      \rowcolor{lightblue}
      \method{}(GRPO)           & 0.92        & 0.88     & 0.77     & 0.94   & 0.46     & 0.72         & 0.79              \\
      \midrule
      Janus & 0.93        & 0.93     & 0.92     & 0.97   & 0.38     & 0.74        & 0.84              \\
      \rowcolor{lightblue}
      \method{}(GRPO)           & 0.94        & 0.95     & 0.83     & 0.99   & 0.52     & 0.89         & 0.88              \\
      \bottomrule
    \end{tabular}%
  }
  \vspace{-15pt}
\end{wraptable}

%% file: tables/mmu_t2i.tex
\begin{table*}
  \centering
\caption{Comparison using our proposed metric to evaluate the imbalance between the two tasks.}
  \label{tab:mmu_t2i}
  \setlength{\tabcolsep}{1.8pt}  
  \resizebox{\linewidth}{!}{%
    \begin{tabular}{lcccccc|c||cccccc|c}
      \toprule
      \multirow{2}{*}{Method}
        & \multicolumn{7}{c||}{$\text{Accuracy}_{\text{MMU}|\text{T2I}}$}
        & \multicolumn{7}{c}{$\text{Accuracy}_{\text{T2I}|\text{MMU}}$} \\
      \cmidrule(r){2-8}\cmidrule(l){9-15}
        & Single. & Two. & Count. & Colors & Posi. & Attri. & Overall$\uparrow$
        & Single. & Two. & Count. & Colors & Posi. & Attri. & Overall$\uparrow$ \\
      \midrule
      Show-o
        & 0.98 & 0.89 & 0.92 & 0.94 & 0.10 & 0.61 & 0.83
        & 0.96 & 0.66 & 0.67 & 0.85 & 0.18 & 0.41 & 0.70 \\
      HermesFlow
        & 0.99 & 0.85 & 0.90 & 0.96 & 0.23 & 0.57 & 0.83
        & 0.97 & 0.68 & 0.69 & 0.84 & 0.24 & 0.41 & 0.71 \\
      \method{}(SFT)
        & 0.11 & 0.99 & 0.97 & 0.99 & 0.83 & 0.98 & 0.81
        & 0.99 & 0.95 & 0.58 & 0.93 & 0.60 & 0.74 & 0.80 \\
      \rowcolor{lightblue}
      \method{}(GRPO)
        & 0.99 & 0.97 & 0.90 & 0.99 & 0.81 & 0.85 & 0.93
        & 0.97 & 0.85 & 0.70 & 0.90 & 0.67 & 0.68 & 0.81 \\
      \midrule

      Janus
        & 0.91 & 0.59 & 0.71 & 0.96 & 0.20 & 0.42 & 0.67
        & 0.95 & 0.67 & 0.27 & 0.90 & 0.39 & 0.49 & 0.68 \\
      \rowcolor{lightblue}
      \method{}(GRPO)
        & 0.94 & 0.78 & 0.52 & 0.98 & 0.31 & 0.77 & 0.76
        & 0.96 & 0.76 & 0.34 & 0.88 & 0.63 & 0.59 & 0.74 \\
      \bottomrule
    \end{tabular}%
  }
  \vspace{-15pt}
 \end{table*}

%% file: sections/conclusion.tex
\section{Conclusion}

In this work, we propose a self-improving post-training method for unified multimodal models. By combining supervised fine-tuning (SFT) with Group Relative Policy Optimization (GRPO), our approach improves both generation and understanding performance while reducing task imbalance. Experiments on Show-o and Janus validate the effectiveness of our method.

%% file: sup.tex
\section{Technical appendices}

In this appendix, we provide additional implementation details of our experiments (Sec.~\ref{more_implementation_detail}) and describe the formulation and explanation of the Straight-Through Gumbel-Softmax estimator (Sec.~\ref{gumbel_softmax_formula}). We then present more details and the loss formulation for the non-end-to-end training pipeline (Sec.~\ref{non_end_to_end}). Additional results for both tasks, along with extended visualization examples, are provided in Sec.~\ref{more_mmu_result}, Sec.~\ref{geneval_more}, and Sec.~\ref{ood_result}. We also offer further analysis comparing SFT and GRPO in non-end-to-end settings (Sec.~\ref{analysis_non_end_to_end_sft}). Finally, we discuss the limitations of our work and its broader impact (Sec.~\ref{limintations} and Sec.~\ref{broader_impact}).

\subsection{More implementation details}
\label{more_implementation_detail}

\textbf{Training details.}
For the SFT or GRPO phase, we set the learning rate to \(1 \times 10^{-5}\) for both the Janus and Show-o models. We use the constructed prompts as input and randomly select one sentence from the six predefined categories. The constructed prompts serve as the training set, while the prompts from the GenEval benchmark are used as the test set.

\textbf{Understanding benchmark.} For the understanding benchmarks, we use other unified multimodal models to generate images based on the prompts from the GenEval benchmark. We then filter and retain only the correctly generated images. These images, along with the constructed questions, are used as inputs to evaluate the model’s understanding capability. We compute the accuracy as the percentage of predicted answers that match the ground-truth answers.

\textbf{New metric.} For the T2I$\rightarrow$MMU chain, we use the prompts from the GenEval benchmark for evaluation. For the MMU$\rightarrow$T2I chain, we follow the input format of our understanding benchmarks, feeding an external image and corresponding questions for evaluation. To assess the imbalance between the two tasks using our proposed metric, we employ external language models such as Qwen~\cite{qwen} or GPT~\cite{gpt_4o} to judge whether the predicted answers match the ground-truth answers and compute accuracy for each category. For image evaluation, we follow the GenEval protocol, using MMDetection~\cite{mmdetection} and Mask2Former~\cite{mask2former} to determine whether the generated image is correct.

\subsection{Straight-Through gumbel softmax}
\label{gumbel_softmax_formula}

The Straight-Through Gumbel-Softmax (ST-GS) estimator enables differentiable sampling of discrete tokens, which is crucial for gradient-based optimization in end-to-end training. It approximates the sampling process using a soft relaxation during the backward pass while retaining hard discrete choices during the forward pass.

The Gumbel-Softmax distribution samples a continuous approximation of one-hot vectors using the following formula:
\begin{equation}
y_i = \frac{\exp\left( \dfrac{\log(\pi_i) + g_i}{\tau} \right)}{\sum\limits_{j=1}^{K} \exp\left( \dfrac{\log(\pi_j) + g_j}{\tau} \right)},
\end{equation}
where \(\pi_i\) is the unnormalized probability of the \(i\)-th category, \(g_i\) is sampled from the \(\text{Gumbel}(0,1)\) distribution, \(\tau\) is a temperature parameter controlling the smoothness, and \(K\) is the number of categories.

In the straight-through version, a hard one-hot vector \(\mathbf{z}\) is obtained during the forward pass by applying \(\arg\max\) over the sampled logits:
\begin{equation}
z_i = \begin{cases}
1, & \text{if } i = \arg\max\limits_j \, y_j \\
0, & \text{otherwise}
\end{cases}.
\end{equation}
While the forward pass uses hard selection, the backward pass uses the soft \(y_i\) values to compute gradients, enabling smooth updates.

We adopt the Straight-Through Gumbel-Softmax in our method to allow gradients from downstream tasks to propagate through discrete image tokens. This is critical for end-to-end training, as it ensures that the generation module receives meaningful gradient signals and participates in joint optimization with the understanding module.

\subsection{Non end-to-end training pipeline}
\label{non_end_to_end}

For non-end-to-end training, we decouple the GRPO loss into two separate objectives for the T2I and MMU modules.

\textbf{T2I optimization.}  
For the image generation task, we directly optimize the image token logits using rewards computed from MMU performance. The GRPO loss is defined as:
\[
\mathcal{L}_{\text{T2I}}(\theta) =
-\,\sum_{k=1}^{K}
w_k\,\log p_\theta\bigl(\hat{\mathbf{u}}_k \mid p\bigr)
+
\beta\,
\mathrm{KL}\left(p_\theta(\mathbf{u} \mid p)\,\|\,p_{\theta_{\mathrm{ref}}}(\mathbf{u} \mid p)\right),
\]
where \(p\) is the input prompt, \(\hat{\mathbf{u}}_k\) is the \(k\)-th sampled image token sequence, and \(w_k\) is a reward-based importance weight derived from MMU outputs.

\textbf{MMU optimization.}  
For the multimodal understanding task, we optimize the answer prediction directly using rewards based on answer correctness. The GRPO loss is given by:
\[
\mathcal{L}_{\text{MMU}}(\theta) =
-\,\sum_{k=1}^{K}
w_k\,\log p_\theta\bigl(\hat{a}_k \mid \mathbf{u},\,q\bigr)
+
\beta\,
\mathrm{KL}\left(p_\theta(a \mid \mathbf{u}, q)\,\|\,p_{\theta_{\mathrm{ref}}}(a \mid \mathbf{u}, q)\right),
\]
where \(\mathbf{u}\) is the input image token sequence, \(q\) is the question, \(\hat{a}_k\) is the predicted answer, and \(w_k\) is computed based on the answer reward.

Although the two modules are optimized independently, reward signals still facilitate cross-task influence, allowing the model to benefit from mutual supervision even without gradient sharing.

\subsection{More generation results}
\label{geneval_more}
In our main experiments using the Show-o model~\cite{showo}, both our method and the baselines adopt a guidance scale of 2 and 16 generation steps during training and inference. Here, we present additional comparison results under a higher guidance scale of 5 and an extended generation length of 50 steps. We observe that our SFT-based method achieves a final score of 0.79, while our GRPO-based method reaches 0.74—both significantly outperforming all baselines.

\input{tables/geneval_guidance}

\begin{wraptable}{r}{0.4\textwidth}
  \vspace{-10pt}
  \centering
  \caption{Benchmark results for multimodal understanding tasks.}
  \label{tab:mmu_benchmark}
  \resizebox{\linewidth}{!}{%
    \begin{tabular}{l|cc}
      \toprule
      Method & POPE$\uparrow$ & MMMU\_val$\uparrow$ \\
      \midrule
      Show-o & 79.8 & 26.7 \\
      \method{} (GRPO) & 78.1 & 25.9 \\
      \bottomrule
    \end{tabular}%
  }
\end{wraptable}

\subsection{More understanding results}
\label{more_mmu_result}
Since our work primarily focuses on the understanding of basic visual attributes, the main results on tasks such as counting and color recognition have been reported in Sec.\ref{quantitative_results}. We further evaluate our method on more comprehensive benchmarks, including POPE\cite{pope} and MMMU~\cite{mmmu}, as shown in Tab.~\ref{tab:mmu_benchmark}. The slightly lower scores can be attributed to the fact that our training data primarily focuses on basic understanding features. Besides, some benchmarks like MMMU focus more on text-based reasoning and numerical calculation, rather than visual understanding. We believe that incorporating a broader range of understanding categories during training, or using a larger model, may help mitigate this limitation in future work.

\begin{figure*}[t]
  \centering
  \includegraphics[width=\linewidth]{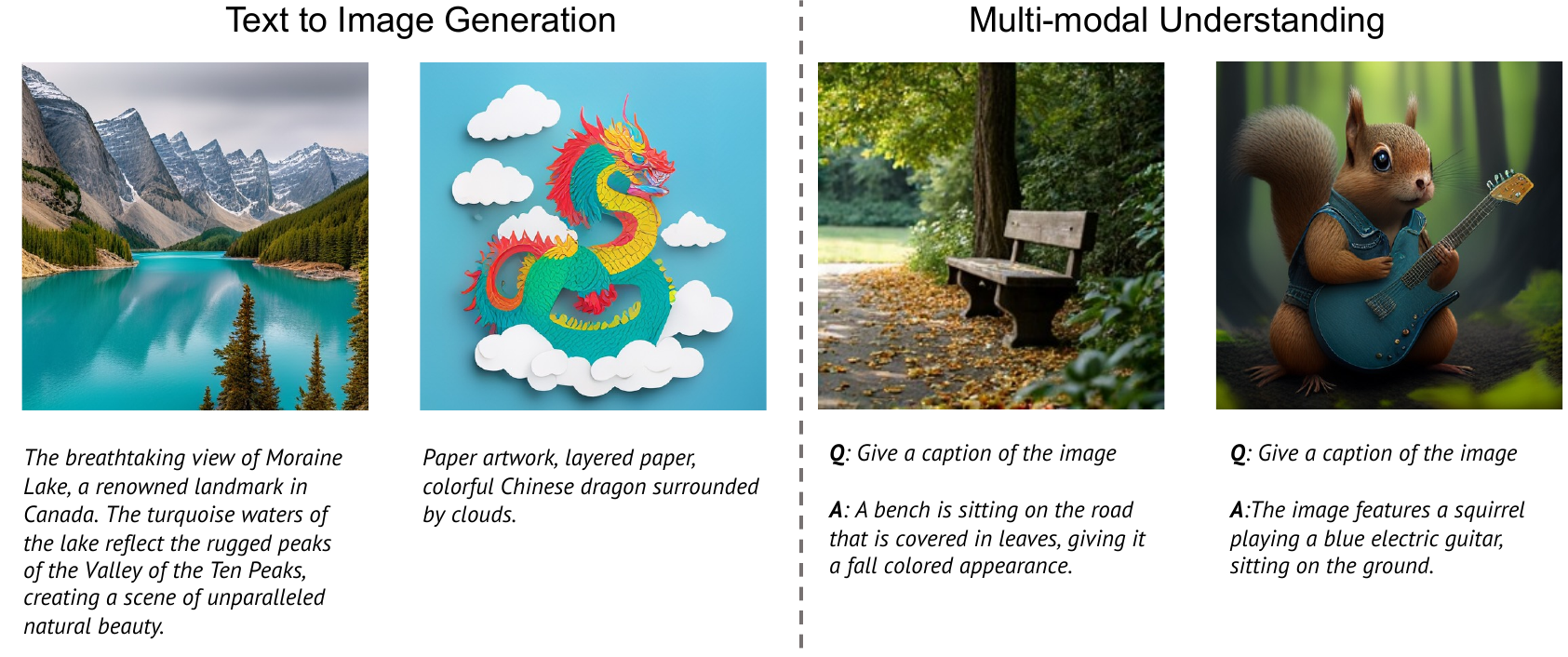}
  \vspace{-10pt}
  \caption{Visualization results of our method.}
  \label{fig:ood}
  \vspace{-10pt}
\end{figure*}

\subsection{More visualization results}
\label{ood_result}

We provide additional visualization results in Fig.~\ref{fig:ood}, including several out-of-distribution examples. These results demonstrate that our method (GRPO) generalizes beyond the specific categories seen during training.

\subsection{More analysis of SFT and GRPO}
\label{analysis_non_end_to_end_sft}

\textbf{End-to-end training.}
For end-to-end training, supervision for T2I is inherently indirect, as the optimization signal comes from downstream tasks such as answer prediction rather than direct supervision of the generated image. Compared to MMU, T2I is generally less prone to overfitting due to the structured nature of image reconstruction. In this setting, SFT may outperform GRPO, as it leverages the ground-truth answer to directly compute the loss and guide the model toward a clearer optimization direction. This process does not heavily depend on the capacity of the base model; instead, the ground-truth serves as a strong supervision signal that helps SFT converge more efficiently.

In contrast, GRPO optimizes the model by generating a group of candidate answers and computing rewards to update the policy. When the model capacity is limited, the generated answers may all be incorrect, leading to weak or noisy reward signals and thus making optimization more difficult.

For MMU, which primarily relies on learning from text-based inputs, SFT tends to overfit easily by memorizing token patterns. GRPO, on the other hand, benefits from its reward-based training and KL regularization term, which encourages the updated model to remain close to the original pretrained model. This constraint improves generalization and helps prevent overfitting, especially in cases where training data is limited or highly structured.

\textbf{Non end-to-end training.}
For non end-to-end training, applying SFT to T2I is often unstable due to error accumulation from the generation pipeline. The optimization direction can easily become unreliable, leading to training collapse. To mitigate this, one option is to follow prior works~\cite{hermesflow,showo_cot}, which use offline-generated images and filter out incorrect samples to construct a reliable training set for SFT. In comparison, GRPO is more robust in non end-to-end settings, as it directly optimizes with reward signals and demonstrates better stability and performance.

\subsection{Limitations}
\label{limintations}
While our work presents an effective post-training method for unified multimodal models, it also has several limitations.
First, our current focus is limited to basic visual understanding tasks such as counting and color recognition. More complex reasoning abilities, such as mathematical problem-solving or abstract inference, are beyond the scope of this work and are left for future exploration. Second, the training speed per iteration is relatively slow. This is a known limitation of GRPO-based methods, as they require multiple forward passes during inference before optimization.

\subsection{Broader impacts}
\label{broader_impact}
Our work explores an effective post-training method for unified multimodal models. We believe our method will not bring negative societal impacts.

%% file: tables/geneval_guidance.tex
\begin{table*}[h]
  \centering
    \caption{Comparison of \method{} with baseline methods on the GenEval benchmark under a guidance scale of 5 and 50 generation steps.}
  \label{tab:t2i_guidance}
  \resizebox{0.8\textwidth}{!}{%
    \begin{tabular}{lcccccc|c}
      \toprule
      Method           & Single. & Two. & Count. & Colors & Posi. & Attri. & Overall$\uparrow$ \\
      \midrule
      Show-o   & 0.98        & 0.81     & 0.69     & 0.82   & 0.32     & 0.53         & 0.68              \\
      HermesFlow & 0.98        & 0.84     & 0.66     & 0.82   & 0.32     & 0.52         & 0.69              \\
      \method{}(SFT)
      & 1.00        & 0.97     & 0.61     & 0.91   & 0.56     & 0.70        & 0.79              \\
      \rowcolor{lightblue}
      \method{}(GRPO)           & 0.96        & 0.80     & 0.67     & 0.86   & 0.50     & 0.67         & 0.74              \\
      \bottomrule
    \end{tabular}%
  }
\end{table*}